\documentclass[11pt]{article}
\usepackage{acl}
\usepackage{times}
\usepackage{pdflscape}
\usepackage{pifont}           
\usepackage{latexsym}
\usepackage[T1]{fontenc}
\usepackage[utf8]{inputenc}
\usepackage{microtype}
\usepackage{inconsolata}
\usepackage{amsmath}
\usepackage{amssymb} 
\usepackage{graphicx} 
\usepackage{adjustbox}
\usepackage{array}
\usepackage[table]{xcolor}
\usepackage{tikz}
\usepackage{placeins}
\usepackage{float}
\usetikzlibrary{positioning}
\definecolor{headergray}{gray}{0.9}  
\usepackage{soul} 
\usepackage{longtable}
\usepackage{makecell}
\usepackage{multirow}
\usepackage{booktabs}
\usepackage{adjustbox}
\newcommand{\symfilled}{$\bullet$}
\newcommand{\symempty}{$\circ$}
\newcommand{\symhybrid}{$\odot$}
\title{AI Alignment through a Game-theoretic Lens: A Survey}

\author{
 \textbf{Yanan Cai\textsuperscript{1,2}}\thanks{Equal contribution.},
 \textbf{Zhongrui Zhao\textsuperscript{1,2}}\footnotemark[1],
 \textbf{Zhigang Lu\textsuperscript{2}}\thanks{Corresponding author.},
 \textbf{Ickjai Lee\textsuperscript{1}},
 \textbf{Wei~Emma~Zhang\textsuperscript{3}},\\
 \textbf{Minhui Xue\textsuperscript{3,4}},
 \textbf{Yihong Zhang\textsuperscript{5}},
 \textbf{Shuchao Pang\textsuperscript{6,7}},
 \textbf{Wei Xiang\textsuperscript{8}}
\\
 \textsuperscript{1}James Cook University,
 \textsuperscript{2}Western Sydney University,
 \textsuperscript{3}Adelaide University,
 \textsuperscript{4}CSIRO,\\
 \textsuperscript{5}The University of Osaka, 
 \textsuperscript{6}Nanjing University of Science and Technology,\\
 \textsuperscript{7}Macquarie University,
 \textsuperscript{8}La Trobe University
\\
   \{yanan.cai, zhongrui.zhao\}@my.jcu.edu.au, z.lu@westernsydney.edu.au
}

\begin{document}
\maketitle

\begin{abstract}
As large language models and increasingly capable AI agents are deployed in high-risk settings, aligning them with complex human values has become a central challenge. Existing alignment methods, while effective in improving helpfulness, harmlessness, and controllability, often struggle to capture real-world preferences that are context-dependent, non-transitive, and shaped by dynamic multi-party interactions. This survey reviews AI alignment through a game-theoretic lens. Specifically, it organizes recent progress around key game-theoretic elements and synthesizes the literature along three challenges: preference diversity, alignment priority, and temporal dynamics. This perspective clarifies where current alignment methods genuinely benefit from game-theoretic analysis, where the framework is looser, and what challenges remain in building robust, adaptive, and verifiable AI systems.
\end{abstract}

\begin{figure}[t!]
\centering 
\includegraphics[width=0.5\textwidth]{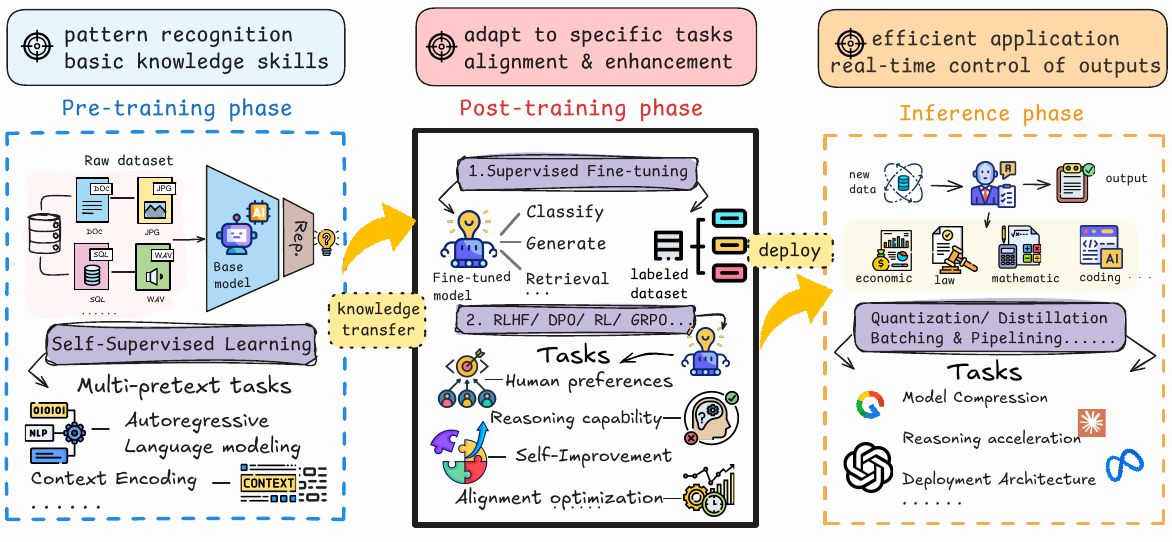}
\includegraphics[width=0.5\textwidth]{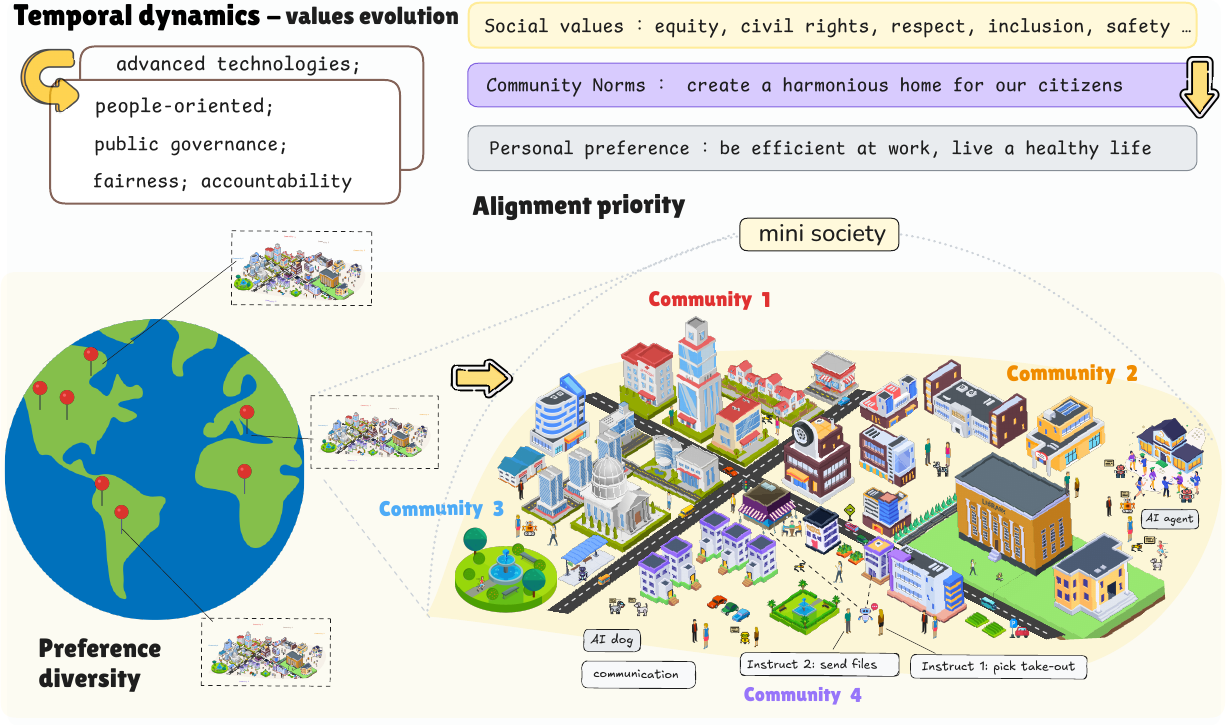}
\caption{Alignment process (top panel) and three core challenges (bottom): preference diversity, alignment priority, and
temporal dynamics.}
\label{fig:post-training}
\end{figure}

\section{Introduction}
\label{sec:introduction}
As large language models (LLMs) and AI agents are increasingly deployed in real-world settings~\cite{xi2025rise}, security and governance risks raised by LLMs and AI agents have received growing attention~\cite{duenas2024frontier, milano2024advanced}. Reported concerns include bias, toxicity, manipulation, and misrepresentation~\cite{duan2025gender, gallegos2024bias, guo2024online}, as well as deception, reward hacking, and power-seeking behavior under imperfect oversight~\cite{pan2022effects, park2024ai, turner2022parametrically}. AI alignment is therefore an urgent goal. It aims to ensure that AI systems not only perform tasks effectively, but also act consistently with human intentions, social norms, and ethical constraints~\cite{shen2025position}. 

While alignment can be applied during training or inference, this survey mainly focuses on the post-training phase in Fig.~\ref{fig:post-training}, which adapts foundation models to human preferences or deployment needs without retraining them from scratch, while encoding behavioral changes into model parameters to avoid repeated inference-time intervention~\cite{lai2025survey}. Methods such as reinforcement learning from human feedback (RLHF) and direct preference optimization (DPO) have improved model helpfulness, harmlessness, and controllability~\cite{ouyang2022training, rafailov2023direct}. However, they still struggle when proxy rewards differ from intended goals, parties involved in the alignment process have conflicting preferences (e.g., Different users may prefer either directness or harm and stereotype avoidance when addressing controversial social questions), or static objectives fail under repeated interaction and environmental changes~\cite{munos2024nash, chu2025stackelberg, tang2025game}. We group the limitations into \emph{three core challenges: preference diversity, alignment priority, and temporal dynamics}.

The core challenges mentioned above motivate a promising and more structural perspective on alignment via game theory~\cite{ye2024online}. Specifically, \emph{preference diversity} arises since values differ across human groups, e.g., ethnic or political groups, making a single tractable and socially legitimate objective hard to define~\cite{gabriel2020artificial, floridi2022unified}; \emph{alignment priority} reflects multiple levels of alignment, including reward proxies, optimization targets, user preferences, and higher legal or institutional constraints, which may conflict~\cite{amodei2016concrete, kleinberg2016inherent}; furthermore, \emph{temporal dynamics} need to reflect the evolution of human values, social environment, and AI capabilities~\cite{carroll2024ai, qiu2024progressgym}. In this sense, we aim to use the game-theoretic language to characterize and explain a series of application scenarios and uncertainties in the alignment problems by analyzing some game elements such as players, utilities, information, strategies, equilibrium, and dynamics~\cite{gemp2024states, mozikov2024eai}. To avoid overstating the role of game theory, we also clarify what game-theoretic analysis can formally guarantee and under which assumptions, rather than claiming that game theory universally solves alignment. In summary, we give the main contributions of this survey below:



\begin{itemize}
    \item We organize a three-dimensional analysis framework of AI alignment by \emph{challenge one - preference diversity}, \emph{challenge two - alignment priority}, and \emph{challenge two - temporal dynamics} for helping researchers clarify the current alignment landscape. 

    \item We provide a unified game-theoretic perspective for AI alignment, which is used to interpret and formalize specific alignment goals and scenarios, via the mathematical foundations and theoretical results of game theory.

    \item We summarize six alignment directions under the challenge: general preference or heterogeneous preference, synchronous interaction or sequential interaction, and adaptive evolution or co-evolution, encouraging new insights within the AI alignment community.
    
\end{itemize}

\section{Related Work}
Some existing surveys have reviewed AI alignment broadly, emphasizing objectives, training pipelines, and governance~\cite{ji2023ai, sun2026towards, shen2025position, kayabay2025data}, while others focus on LLM agents and multi-agent systems, organizing the literature around architectures, workflows, and deployment challenges~\cite{wang2024survey, li2024survey, zhou2025survey}. These works provide a foundational understanding of AI alignment for us. Recently, researchers have started to apply the game-theoretic perspective to apply game theory perspectives to various aspects of alignment, such as preference conflict, equilibrium-based alignment, cooperative AI, online adaptation, and pluralistic evaluation~\cite{tang2025game, chu2025stackelberg, ye2024online, mao2025alympics}. However, these efforts remain scattered across different scene settings and research communities, lacking a unified organization. Sun et al.~\cite{sun2025game} highlight the broad intersection between language models and games in benchmarking, strategic behavior, and social simulation, but mainly focus on examining the bidirectional relationship between the two. Hence, our survey, focusing on post-training alignment approaches, asks how game theory determines the formulation, guarantee, or evaluation of alignment problems. Appx.~\ref{appx:details} and \ref{appx:benchmark} give method details and evaluation benchmarks, respectively.

%

\begin{figure*}[t!]
\begin{center}
\begin{adjustbox}{center,scale=0.45} 
\begin{tikzpicture}[
  font=\Large,
  box/.style={
    fill=white!95!yellow!,
    rounded corners,
    align=center,
    minimum width=1cm,
    minimum height=8mm,
    text width=8cm
  },
  box11/.style={
    draw=pink!90!black,
    line width=2pt,
    rounded corners,
    align=center,
    minimum width=1cm,
    minimum height=10mm,
    text width=4.8cm
  },
  box12/.style={
    draw=pink!90!black,
    line width=2pt,
    rounded corners,
    align=center,
    minimum width=3cm,
    minimum height=7mm,
    text width=8.5cm
  },
  box13/.style={
    draw=pink!90!black,
    line width=2pt,
    rounded corners,
    align=center,
    minimum width=3cm,
    minimum height=7mm,
    text width=3.8cm
  },
   box14/.style={
    line width=2pt,
    fill=pink!90,
    rounded corners,
    align=center,
    minimum width=1cm,
    minimum height=9mm,
    text width=12cm,
    font=\normalsize,
  },
  box21/.style={
    draw=white!50!cyan!,
    rounded corners,
    line width=2pt,
    align=center,
    minimum width=1cm,
    minimum height=10mm,
    text width=4.9cm,  
  },
  box22/.style={
    draw=white!50!cyan!,
    line width=2pt,
    rounded corners,
    align=center,
    minimum width=1cm,
    minimum height=7mm,
    text width=8.5cm 
  },
    box23/.style={
    line width=2pt,
    rounded corners,
    fill=white!80!cyan!,
    align=center,
    minimum width=2cm,
    minimum height=9mm,
    text width=12cm,
    font=\normalsize
  },
  box31/.style={
    draw=brown!,
    rounded corners,
    line width=2pt,
    align=center,
    minimum width=1cm,
    minimum height=10mm,
    text width=4.8cm,    
  },
  box32/.style={
    draw=brown!,,
    line width=2pt,
    rounded corners,
    align=center,
    minimum width=3cm,
    minimum height=7mm,
    text width=3.8cm,  
  },
    box33/.style={
    draw=brown!,
    line width=2pt,
    rounded corners,
    align=center,
    minimum width=3cm,
    minimum height=7mm,
    text width=8.5cm,  
  },
   box34/.style={
    fill=brown!60,
    line width=2pt,
    rounded corners,
    align=center,
    minimum width=1cm,
    minimum height=9mm,
    text width=12cm,  
    font=\normalsize
  },
 box41/.style={
    draw=white!20!gray!,
    rounded corners,
    line width=2pt,
    align=center,
    minimum width=1cm,
    minimum height=10mm,
    text width=4.8cm,  
  },
  box42/.style={
    draw=white!20!gray!,
    line width=2pt,
    rounded corners,
    align=center,
    minimum width=3cm,
    minimum height=5.5mm,
    text width=3.8cm, 
  },
    box43/.style={
    line width=2pt,
    rounded corners,
    fill=white!80!gray!,
    align=center,
    minimum width=2cm,
    minimum height=9mm,
    text width=12cm,
  },
   box44/.style={
    draw=white!20!gray!,
    line width=2pt,
    rounded corners,
    align=center,
    minimum width=3cm,
    minimum height=6mm,
    text width=8.5cm,
  },
  arr/.style={-}
]
\node[rotate=90, transform shape, box, opacity=0] (rootbase)
{\textbf{Alignment meets game theory}};

\node[draw, rotate=90, transform shape, box]
(root) at ([yshift=6.7mm]rootbase.center)
{\textbf{Alignment meets game theory}};

\draw[arr] (root.south) -- ++(6mm,0) coordinate(l0);


\node[box11, above right=65mm and 20mm of rootbase](lrss){\textbf{Challenge One}: \\Preference diversity \\(Sec.~\ref{sec:challenge_1})};
\draw[arr](l0) |- (lrss.west);

\node[box11, above right=28mm and 6mm of lrss](gp){General preference};
\node[box11, below right=23mm and 6mm of lrss](hp){Heterogeneous preference};

\draw[arr](lrss.east)--++(4mm, 0) coordinate(l1);
\draw[arr](l1) |- (hp.west);
\draw[arr](l1) |- (gp.west);

\node[box12, above right=5.5mm and 6mm of gp](nb){Non-game: best response toward a fixed policy};
\node[box13, below right=6mm and 6mm of gp](ft){Two-player game};
\draw[arr](gp.east)--++(4mm, 0) coordinate(l2);
\draw[arr](l2) |- (nb.west);
\draw[arr](l2) |- (ft.west);

\node[box13, above right=5mm and 6mm of hp](ng){Non-game};
\node[box12, below right=3mm and 6mm of hp](mpg){Multi-player game: preference aggregation \& consensus};
\draw[arr](hp.east)--++(4mm, 0) coordinate(l2);
\draw[arr](l2) |- (ng.west);
\draw[arr](l2) |- (mpg.west);

\node[box13, above right=7mm and 6mm of ft](ai){Existence\\convergence};
\node[box13, right=3mm and 6mm of ft](ec){Average-iterate\\convergence};
\node[box13, below right=8mm and 6mm of ft](li){Last-iterate\\convergence};

\node[box13, above right=0.5mm and 6mm of ng](sfo){Social welfare optimization };
\node[box13, below right=0.5mm and 6mm of ng](mbo){Multi-objective \\optimization};

\draw[arr](ft.east) --++ (4mm, 0) coordinate(l4);
\draw[arr](l4) |- (ai.west);
\draw[arr](l4) -- (ec.west);
\draw[arr](l4) |- (li.west);

\draw[arr](ng.east) --++ (4mm, 0) coordinate(l4);
\draw[arr](l4) |- (sfo.west);
\draw[arr](l4) |- (mbo.west);

\node[box14, right=0mm and 6mm of nb](f7){IPO~\cite{azar2024general}};
\node[box14, right=0mm and 6mm of ai](f1){Online Iterative RLHF~\cite{ye2024online}};

\node[box14, right=0mm and 6mm of ec](f2){SPO\cite{swamy2024minimaximalist}, DNO~\cite{rosset2024direct}, REBEL~\cite{gao2024rebel}, SPPO~\cite{wu2025self}, TANPO~\cite{wangprovably}, ONPO~\cite{zhang2025improving}};

\node[box14, right=0mm and 6mm of li](f3){NLHF~\cite{munos2024nash}, IPO-MD~\cite{calandriello2024human}, INPO~\cite{zhangiterative}, COMAL~\cite{liu2026comal},  MPO~\cite{wang2025magnetic}, RSPO~\cite{tang2025game}};

\draw[arr](nb.east) -- (f7.west);
\draw[arr](ai.east) -- (f1.west);
\draw[arr](ec.east) -- (f2.west);
\draw[arr](li.east) -- (f3.west);

\node[box14, right=0mm and 6mm of sfo](f4){~\cite{chidambaram2026direct}, ~\cite{kim2026beyond}, MOP~\cite{xiongprojection}, Pessimistic Nash Bargaining~\cite{zhong2024provable}, GRPO~\cite{ ramesh2024group}};

\node[box14, right=0mm and 6mm of mbo](f5){CLP~\cite{wang2024conditional}, Panacea~\cite{zhong2024panacea}, MO-ODPO~\cite{gupta2025robust}, SIPO~\cite{li2025self}, DPA~\cite{wang2024arithmetic}, DRM~\cite{luo2025rethinking}};

\node[box14, right=0mm and 6mm of mpg](f6){RLHF Game~\cite{sun2025mechanism}, Dynamic Bayesian Game Formulation~\cite{hao2025online}, Battling Influencers Game~\cite{wu2025battling}, 〈M, N , $\epsilon$, $\delta$〉-Agreement\cite{nayebi2026intrinsic}};

\draw[arr](sfo.east) -- (f4.west);
\draw[arr](mbo.east) -- (f5.west);
\draw[arr](mpg.east) -- (f6.west);

\node[box21, below right=25mm and 11mm of rootbase](piat){\textbf{Challenge Two}: \\Alignment priority \\(Sec.~\ref{sec:challenge_2})};
\draw[arr](l0) |- (piat.west);
\node[box21, above right=5mm and 6mm of piat](si){Synchronous interaction};
\node[box21, below right=5mm and 6mm of piat](sis){Sequential interaction};

\draw[arr](piat.east)--++(4mm,0) coordinate(l1);
\draw[arr](l1) |- (si.west);
\draw[arr](l1) |- (sis.west);

\node[box22, above right=1mm and 6mm of si](sd){Attacker-defender: self-generated data
};
\node[box22, below right=1mm and 6mm of si](ea){Generator-evaluator: self-annotated data 
};

\draw[arr](si.east)--++(4mm,0) coordinate(l2);
\draw[arr](l2) |- (sd.west);
\draw[arr](l2) |- (ea.west);

\node[box22, above right=1mm and 6mm of sis](po){Leader-follower: self-annotated data
};
\node[box22, below right=0.5mm and 6mm of sis](ea2){Orderly cooperation: external annotated data 
};
\draw[arr](sis.east)--++(4mm,0) coordinate(l3);
\draw[arr](l3) |- (po.west);
\draw[arr](l3) |- (ea2.west);

\node[box23, right=0mm and 6mm of sd](g){\cite{jatova2024employing}, GPO~\cite{zheng2025toward}, SELF-REDTEAM~\cite{liu2026chasing}, MTSA\cite{guo2025mtsa}, SPAG~\cite{cheng2024self}};
\node[box23, right=0mm and 6mm of ea](a){APO~\cite{cheng2024adversarial}, SPIN~\cite{chen2024self}, LANA~\cite{azarafrooz2024language}, DuoGuard~\cite{deng2026enhancing}, SPC~\cite{chen2025spc}, PEG~\cite{chen2025incentivizing}};

\draw[arr](sd.east) -- (g.west);
\draw[arr](ea.east) -- (a.west);

\node[box23, right=0mm and 6mm of po](s){STA-RLHF~\cite{makar2024sta}, SPAC~\cite{ji2024self}, SGPO~\cite{chu2025stackelberg}};
\node[box23, right=0mm and 6mm of ea2](sg){ Anyprefer~\cite{zhou2025anyprefer}, CORY~\cite{ma2024coevolving}};

\draw[arr](po.east) -- (s.west);
\draw[arr](ea2.east) -- (sg.west);

\node[box31, below right=120mm and 12mm of rootbase](pdeot){\textbf{Challenge Three}: \\Temporal dynamics \\(Sec.~\ref{sec:challenge_3})};

\draw[arr](l0) |- (pdeot.west);
\node[box31, above right=14mm and 6mm of pdeot](ae){Adaptive evolution };
\node[box31, below right=11mm and 6mm of pdeot](ce){Co-evolution};

\draw[arr](pdeot.east)--++(4mm,0) coordinate(l1);
\draw[arr](l1) |- (ae.west);
\draw[arr](l1) |- (ce.west);

\node[box32, above right=5mm and 6mm of ae](ng){Trajectory-based};
\node[box32, below right=5mm and 6mm of ae](gb){Selection-based};

\draw[arr](ae.east)--++(4mm,0) coordinate(l2);
\draw[arr](l2) |- (ng.west);
\draw[arr](l2) |- (gb.west);

\node[box33, above right=4mm and 6mm of ce](mb){Game-theoretic algorithm modeling
};
\node[box32, below right=2mm and 6mm of ce](sb){Scenario simulation 
};

\node[box32, above right=1mm and 6mm of sb](ss){Social activities
};
\node[box32, below right=1mm and 6mm of sb](gs){Strategic games
};

\draw[arr](sb.east)--++(4mm,0) coordinate(l5);
\draw[arr](l5) |- (ss.west);
\draw[arr](l5) |- (gs.west);

\draw[arr](ce.east)--++(4mm,0) coordinate(l3);
\draw[arr](l3) |- (mb.west);
\draw[arr](l3) |- (sb.west);


\node[box32, right=0mm and 6mm of ng](mdp){Non-MDP / POMDP};
\draw[arr](ng.east) -- (mdp.west);

\node[box32, above right=1mm and 6mm of gb](mg){Evolutionary game 
};
\node[box32, below right=1mm and 6mm of gb](sv){Creator-solver game
};

\draw[arr](gb.east)--++(4mm,0) coordinate(l4);
\draw[arr](l4) |- (mg.west);
\draw[arr](l4) |- (sv.west);


\node[box34, right=0mm and 6mm of mdp](f1){~\cite{klassen2024pluralistic}, ProgressGym~\cite{qiu2024progressgym}};
\draw[arr](mdp.east)-- (f1.west);

\node[box34, right=0mm and 6mm of mg](f2){~\cite{suzuki2024evolutionary}, Evolutionary Agent~\cite{li2024agent}};
\draw[arr](mg.east)-- (f2.west);

\node[box34, right=0mm and 6mm of sv](f3){EVA~\cite{ye2024evolving}};
\draw[arr](sv.east)-- (f3.west);

\node[box34, right=0mm and 6mm of mb](gr){SPARTA\cite{jiang2025sparta}, Red-Team Game~\cite{ma2026evolving}};
\draw[arr](mb.east)-- (gr.west);

\node[box34, right=0mm and 6mm of ss](aly){SSI~\cite{liu2024training}, Artificial Leviathan~\cite{dai2026artificial}, CAMEL~\cite{li2023camel}, AI Collectives~\cite{lai2024position}};
\draw[arr](ss.east)-- (aly.west);

\node[box34, right=0mm and 6mm of gs](gs-ref){~\cite{du2024improving}, Alympics~\cite{mao2025alympics}, FairMindSim~\cite{lei2026llms}, MAD~\cite{liang2024encouraging}};
\draw[arr](gs.east)--(gs-ref.west);

\end{tikzpicture}
\end{adjustbox}
\caption{Taxonomy of alignment based on a game theory lens. }
\label{fig:taxonomy}
\end{center}
\end{figure*}
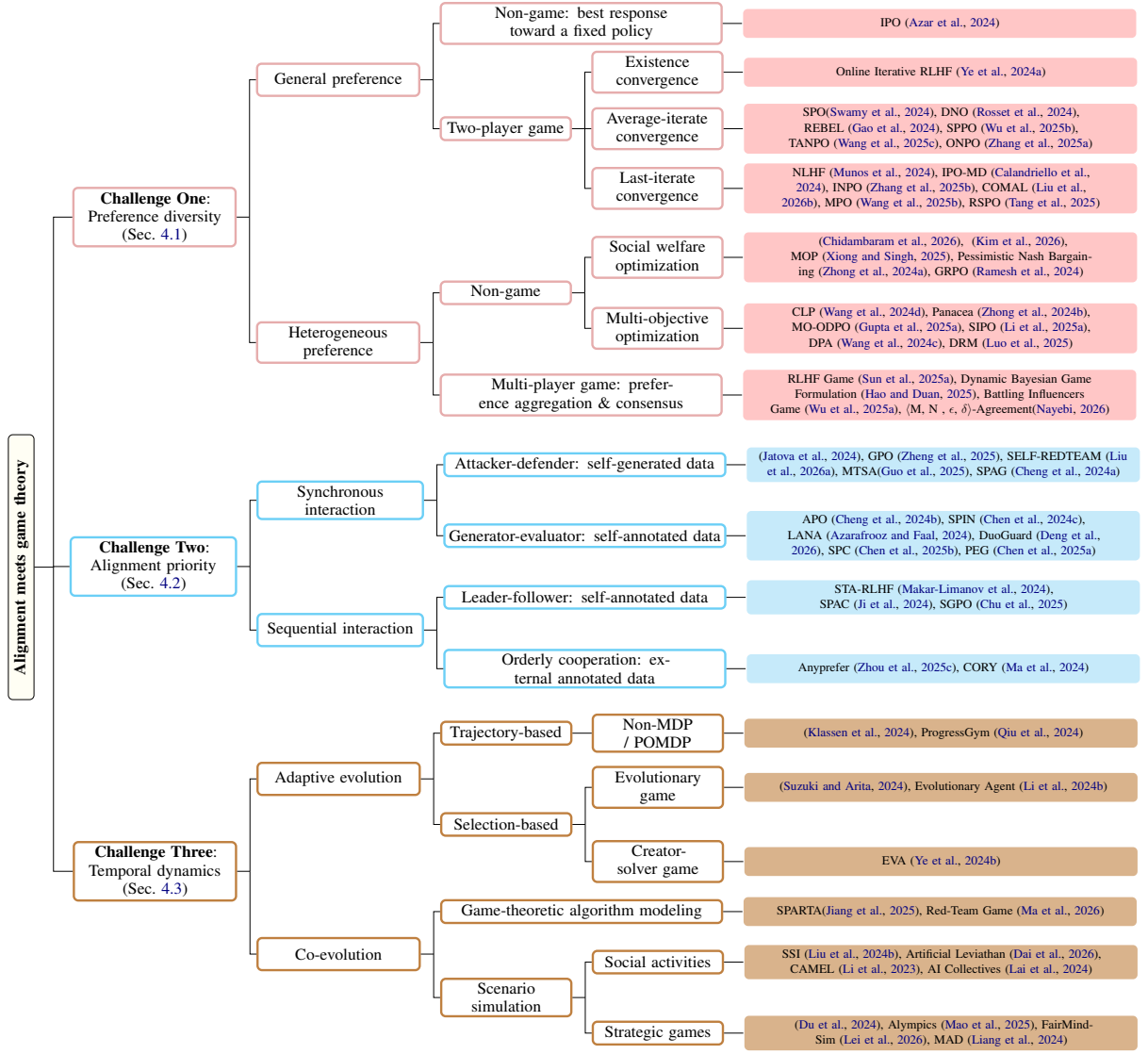

\section{Preliminaries}
In this section, we introduce the preliminaries to understand the role of game theory in AI alignment.

\subsection{General Alignment Guarantee}
Since two-player game theory is the most common formalization of alignment problems, we take its modeling of general preference scenarios as a basis to interpret how to achieve theoretical alignment by pursuing game equilibrium.

\subsubsection{Problem Formulation}
As noted earlier, aligning LLMs typically relies on learning from human preference feedback. Rather than providing a scalar reward signal $r(x,y)$, human feedback more often takes the form of a pairwise comparison signal $\Pr(y_1 \succ y_2 \mid x)$ denoting the probability that $y_1$ is preferred over $y_2$ given prompt $x$, where $\succ$ indicates a preference relation. Accordingly, preference learning more closely matches practical RLHF settings and is better suited to tasks that are difficult to specify using a single numerical reward. In this setting, we seek a policy that performs robustly under human pairwise comparisons against alternative policies. Specifically, consider the following objective
\begin{equation}
\label{eq:obj}
\begin{aligned}
J(\pi_1,\pi_2)
&=
\mathbb{E}_{x \sim d_0}
\mathbb{E}_{y_1 \sim \pi_1,\; y_2 \sim \pi_2}
\Big[
\mathbb{P}(x,y_1,y_2) \\
&\qquad -\tau D_{\mathrm{KL}}\!\bigl(\pi_1(\cdot \mid x)\,\|\,\pi_{ref}(\cdot \mid x)\bigr) \\
&\qquad +\tau D_{\mathrm{KL}}\!\bigl(\pi_2(\cdot \mid x)\,\|\,\pi_{ref}(\cdot \mid x)\bigr)
\Big],
\end{aligned}
\end{equation}
where $d_0$ is a prompt distribution. Here, $\mathbb{P}(x,y_1,y_2)$ is a payoff term instantiated usually as a preference model: $\mathbb{P}(x,y_1,y_2)=\Pr(y_1\succ y_2\mid x)\in[0,1]$ (In many alignment scenarios, the payoff can also be a point reward or other forms). The KL (Kullback–Leibler) terms regularize both policies toward a reference policy ($\pi_{ref}$ denotes a fixed conditional language-model distribution, typically instantiated as the initial supervised fine-tuned model, i.e., $\pi_{ref}=\pi_{SFT}$), thereby stabilizing learning and controlling deviation from the instruction-following behavior~\cite{rafailov2023direct}.

When the payoff term $\mathbb{P}$ satisfies $\Pr(y_1\succ y_2\mid x)+\Pr(y_2\succ y_1\mid x)=1$, the above objective function defines a two-player constant-sum game, i.e., $J(\pi_1,\pi_2)+J(\pi_2,\pi_1)=1$, where the players choose policies $\pi_1$ and $\pi_2$. In particular, $\mathbb{P}$ further can take the centered form of $\mathbb{P}(x,y_1,y_2)=\Pr(y_1\succ y_2\mid x)-\frac{1}{2}$, with a two-player zero-sum game~\cite{liu2026comal}. Based on games, the first player maximizes $J(\pi_1,\pi_2)$, while the second player minimizes it. Intuitively, $\pi_1$ seeks a policy that is preferred under human comparisons, whereas $\pi_2$ acts as an adversarial comparator. 
The alignment problem can therefore be formulated as the saddle-point problem
\begin{equation}
\label{eq:saddle-point}
(\pi_1^{*}, \pi_2^{*}) \in \arg \max_{\pi_1\in\Pi} \arg\min_{\pi_2\in\Pi} J(\pi_1,\pi_2),
\end{equation}
where $\Pi$ is the shared policy class for both players.

\subsubsection{Nash Equilibrium - Strategic Alignment}
For analysis, we restrict attention to policies in $\Pi$ that have the same support as $\pi_{ref}$. Under this assumption, \cite{ye2024online} shows that the game admits a unique Nash equilibrium. By symmetry, the equilibrium policies coincide, i.e., $\pi_1^{*}=\pi_2^{*}=\pi^{*}$. We interpret $\pi^{*}$ as the aligned policy because it satisfies the saddle-point property of the equilibrium. For any alternative policy $\pi \in \Pi$, we have
$
J(\pi,\pi^*) \le J(\pi^*,\pi^*) \le J(\pi^*,\pi).
$
Hence, no policy can systematically obtain a better payoff against $\pi^{*}$ than $\pi^{*}$ obtains against itself, while $\pi^{*}$ performs at least as well against any opponent as it does against itself. Further, when $J(\pi_1,\pi_2)$ is the unregularized objective with a centered pairwise-preference payoff, since $\mathbb{P}(x,y_1,y_2)$ is antisymmetric, we have $J(\pi^\ast,\pi^\ast)=0$. The saddle-point property then gives, for any $\pi\in\Pi$, $J(\pi^\ast,\pi)\geq J(\pi^\ast,\pi^\ast)=0$. Based on the form of $\mathbb{P}(x,y_1,y_2)$, the win rate of $\pi^{*}$ against any $\pi$ is at least one half. Under this view, alignment is not defined as maximizing an unobserved utility function. Rather, it is defined as non-domination under the induced preference relation. In this sense, an aligned policy is strategically stable under pairwise comparison and resistant to exploitation by competing policies.

\subsubsection{Approximation and Convergence}
To assess how well a candidate policy approximates $\pi^{*}$, we use the duality gap~\cite{zhangiterative}:
$
\mathrm{DualGap}(\pi)
:=
\max_{\pi_1 \in \Pi} J(\pi_1,\pi)
-
\min_{\pi_2 \in \Pi} J(\pi,\pi_2).
$
The duality gap is always nonnegative, and $\mathrm{DualGap}(\pi)=0$ if and only if $\pi=\pi^{*}$. Thus, $\mathrm{DualGap}(\pi)$ quantifies how far $\pi$ is from the aligned Nash policy. A small duality gap means that $\pi$ is only weakly exploitable in the induced preference game, providing a formal certificate of approximate alignment. When $\mathrm{DualGap}(\pi) \le \epsilon$, $\pi$ may be regarded as an $\epsilon$-approximate Nash policy. Related definitions and guarantees also appear in~\cite{ye2024online,ma2026evolving}.

The practical significance of this equilibrium formulation depends on whether the learning dynamics (policy updates induced by the alignment procedure itself) converge to such a stable point. Recent works~\cite{munos2024nash,wang2025magnetic,liu2026comal} establish convergence guarantees for several algorithms, including average-iterate convergence of $\frac{1}{T}\sum_{t=1}^{T}\pi_t$. Moreover, if an algorithm enjoys last-iterate convergence, then the final output policy $\pi_{T}$ itself can be interpreted as a strategically aligned solution, and the resulting model inherits the same stability interpretation~\cite{tiapkin2025proximal}. That implies that alignment is not merely observed empirically, but rather the limit of a strategic learning process.

\subsection{Complex Alignment Extensions}

The formulation above can be extended by relaxing its assumptions on the number of players, policy spaces, player roles, action order, and temporal dependence. Multi-party alignment can be modeled as
\(
\mathcal{G}
=
\left\langle
\mathcal{N},
\{\Pi_i\}_{i\in\mathcal{N}},
\{U_i\}_{i\in\mathcal{N}}
\right\rangle,
\)
where $\mathcal{N}=\{1,\ldots,n\}$, $\Pi_i$ is player $i$'s policy space, and $U_i(\pi)$ is its utility under the joint policy
$\pi=(\pi_1,\ldots,\pi_n)$. A profile $\pi^\ast$ is a Nash equilibrium if
\(
U_i(\pi_i^\ast,\pi_{-i}^\ast)
\geq
U_i(\pi_i,\pi_{-i}^\ast),
\forall i\in\mathcal{N},\ \pi_i\in\Pi_i.
\)
Equivalently, each player may minimize
$\ell_i=-U_i$ \cite{wu2025battling}. Non-game extensions instead optimize an aggregated objective
$W(\pi)=F(U_1(\pi),\ldots,U_n(\pi))$, such as social welfare or a multi-objective criterion.

For some asymmetric roles and heterogeneous policy spaces, a general adversarial formulation is
\(
\max_{\pi_1\in\Pi_1}\min_{\pi_2\in\Pi_2}
J(\pi_1,\pi_2),
\)
with Nash equilibrium
\(
J(\pi_1,\pi_2^\ast)
\leq
J(\pi_1^\ast,\pi_2^\ast)
\leq
J(\pi_1^\ast,\pi_2),
\)
where the second condition defines a saddle point
\cite{zheng2025toward,chen2025spc}. When players' actions are sequential, the corresponding Stackelberg equilibrium is
\(
\pi_2^\ast(\pi_1)
\in
\arg\max_{\pi_2\in\Pi_2}J_2(\pi_1,\pi_2),
\pi_1^\ast
\in
\arg\max_{\pi_1\in\Pi_1}
J_1\bigl(\pi_1,\pi_2^\ast(\pi_1)\bigr).
\)
This assumes a selected follower best response
\cite{chu2025stackelberg}; sequential cooperation can instead use a shared-objective Markov game
\cite{zhou2025anyprefer}.

Temporal extensions distinguish adaptations to repeated strategies from changing human values. Over a horizon of $j$ rounds, an $\epsilon$-Nash policy satisfies
\(
U_i^j(\pi_i^\ast,\pi_{-i}^\ast)
\geq
U_i^j(\pi_i,\pi_{-i}^\ast)-\epsilon,
\forall i,\ \pi_i\in\Pi_i,
\)
where $U_i^j$ denotes the accumulated payoff
\cite{ma2026evolving}. Adapting to evolving human values can be modeled by
\(
\mathcal{M}
=
\left\langle
\mathcal{S},\mathcal{A},\mathcal{T},\Omega,O,U
\right\rangle,
U:(\mathcal{S}\times\mathcal{A})^\ast\rightarrow\mathbb{R},
\)
where $\mathcal{S}$ represents latent value states,
$\mathcal{T}$ their dynamics, $(\Omega,O)$ the observations and observation model, and $U$ the trajectory-level alignment utility
\cite{qiu2024progressgym}.

\section{Taxonomy: A Game-theoretic Lens on Alignment Challenges}
\label{sec:alignment_methods}


\subsection{Challenge One: Preference Diversity}
\label{sec:challenge_1}
Preference alignment studies how AI systems are optimized from feedback. In practice, it typically involves three components: the AI system, the feedback source, and a proxy that models the feedback for learning, e.g., a reward or preference model~\cite{ji2023ai}. This section focuses on the structure and diversity of preferences. A game-theoretic lens is particularly useful in two settings: general preference, where robustness is studied through strategic comparison with alternative policies, and heterogeneous preferences, where conflicting party values must be aggregated fairly and stably.
\subsubsection{General Preference}

\begin{figure*}
    \centering
    \includegraphics[width=0.88\linewidth]{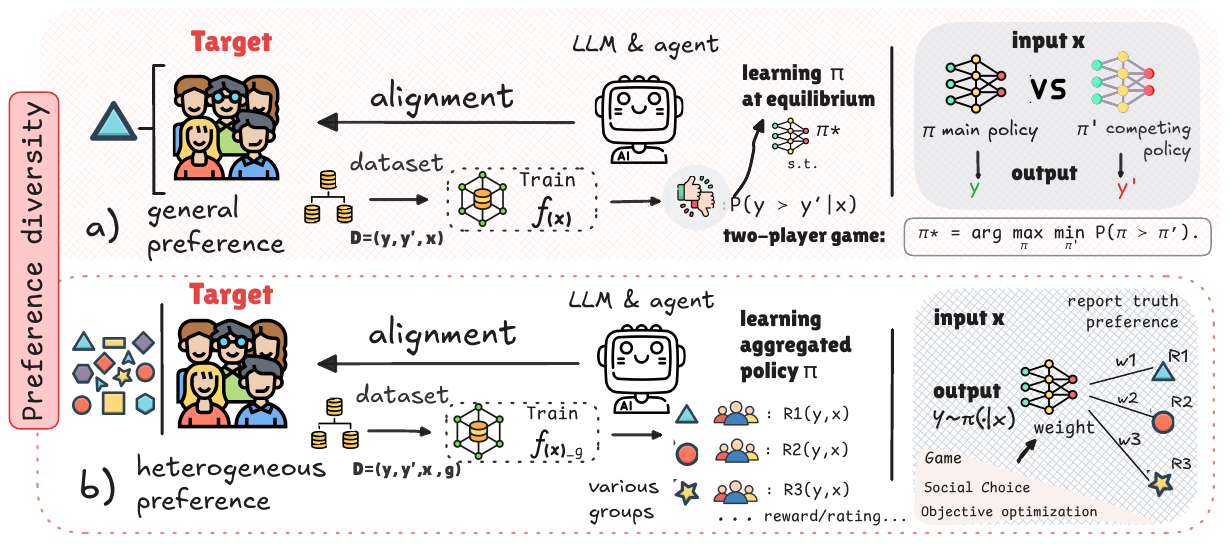}
    \caption{General preference alignment: pairwise optimization and game-theoretic formulation. Heterogeneous preference alignment: multi-group feedback and preference aggregation.}
    \label{fig:preference}
\end{figure*}

\paragraph{Non-game.} General preference alignment aims to optimize model behavior toward broadly shared human judgments, as shown in Fig.~\ref{fig:preference}(a). Early work mainly provided theoretical perspectives on learning from human preferences~\cite{wang2023rlhf}, but did not directly capture practical LLM alignment under pairwise feedback. The initial work is non-game optimization against a fixed policy. The representative example is IPO~\cite{azar2024general}, which directly optimizes pairwise preferences without assuming an explicit reward model:
$
\max_{\pi} \mathbb{E}_{y \sim \pi,\, y' \sim \mu} \left[ \mathbb{P}(y \succ y') \right] - \tau \, \mathrm{KL}(\pi \| \pi_{\mathrm{ref}}).
\label{eq:ipo} 
$
Compared with point-wise reward optimization, IPO better matches preference learning. However, unlike the minimax games, IPO optimizes only the target policy $\pi$ as a best response to a predefined reference distribution $\mu$, rather than jointly optimizing two competing policies. Specifically, $\mu$ plays a role analogous to $\pi_2$ in Eq.~\eqref{eq:saddle-point}, but is fixed in advance and is neither learned nor updated during training. Therefore, IPO does not guarantee robustness against stronger competing policies.

\paragraph{Two-player game.} This limitation motivates a second line of work that formulates alignment as a two-player game. Representative studies such as NLHF~\cite{munos2024nash}, IPO-MD~\cite{calandriello2024human}, and iterative RLHF~\cite{ye2024online} cast alignment as a constant-sum or minimax game, where the target policy and alternative policy are optimized in competition with each other under a learned preference function, updating alternately through self-play in practice. Early results mainly established the existence of Nash-style aligned policies. Subsequent work focused on stronger convergence guarantees. One branch studies average-iterate convergence, including SPO~\cite{swamy2024minimaximalist}, DNO~\cite{rosset2024direct}, REBEL~\cite{gao2024rebel}, SPPO~\cite{wu2025self}, TANPO~\cite{wangprovably}, and ONPO~\cite{zhang2025improving}. These methods often replace explicit win-rate estimation with simpler regression-style objectives or incorporate online exploration to improve sample efficiency. Another branch studies last-iterate convergence, which is more desirable because the final learned policy itself inherits the equilibrium interpretation. Representative examples include INPO~\cite{zhangiterative}, COMAL~\cite{liu2026comal}, MPO~\cite{wang2025magnetic}, and RSPO~\cite{tang2025game}. Overall, general preference alignment has evolved from fixed-opponent optimization to self-play Nash learning, with increasing emphasis on convergence, robustness, and the role of regularization.

\subsubsection{Heterogeneous Preferences}

Heterogeneous preference alignment concerns settings where conflicting parties hold overlapping or conflicting values, as shown in Fig.~\ref{fig:preference}(b). Unlike general preference alignment, the goal is not to fit a single dominant preference, but to balance multi-party, objectives, and minority concerns.

\paragraph{Non-game.} A first category is non-game alignment. One line formulates the problem as social welfare optimization, where group utilities are aggregated through welfare or bargaining objectives~\cite{mishra2023ai,conitzer2024social,qiu2024representative}. Representative methods include MOP~\cite{xiongprojection}, Pessimistic Nash Bargaining~\cite{zhong2024provable}, and group-robust approaches~\cite{chidambaram2026direct,ramesh2024group,kim2026beyond}. These methods improve fairness and reduce dominance, but their behavior depends strongly on the chosen welfare function. Another line formulates alignment as multi-objective optimization. Methods such as CLP~\cite{wang2024conditional}, Panacea~\cite{zhong2024panacea}, MO-ODPO~\cite{gupta2025robust}, and SIPO~\cite{li2025self} jointly optimize multiple objectives and often support controllable trade-offs through preference vectors or conditional signals. Others also study decomposed or vector-valued reward representations for more interpretable preference control~\cite{wang2024arithmetic,luo2025rethinking}.

\paragraph{Multi-player game.} A second category treats heterogeneous alignment as a multi-player game over preference aggregation. Here the main issue is that parties may strategically report preferences to influence the final model. RLHF Game~\cite{sun2025mechanism} and Dynamic Bayesian Game~\cite{hao2025online} explicitly study truthful reporting and weighting mechanisms, while BIG~\cite{wu2025battling} analyzes strategic influence among multiple parties. Other work further examines multi-preference optimization and the complexity of alignment across multiple objectives and agents under weak assumptions~\cite{gupta2025ampo,nayebi2026intrinsic}. Together, these studies show that heterogeneous preference alignment is not only an optimization problem, but also a mechanism-design problem shaped by strategic human behavior.

\subsection{Challenge Two: Alignment Priority}
\label{sec:challenge_2}
This section shifts the focus to interaction structure. AI alignment is increasingly understood as an interactive process rather than a one-shot transfer of human preferences to a model~\cite{shen2025position}. Many methods improve alignment not only by fitting feedback, but also by generating informative feedback through play, critique, or role-dependent adaptation. The main distinction here is whether strategic updates occur within a shared decision round or under an explicit order of moves. We therefore distinguish between synchronous interaction, which is especially useful for adversarial stress testing and self-improvement, and sequential interaction, which introduces commitment, asymmetry, and bilevel optimization into alignment.

\begin{figure*}[t!]
    \centering
    \includegraphics[width=0.88\linewidth]{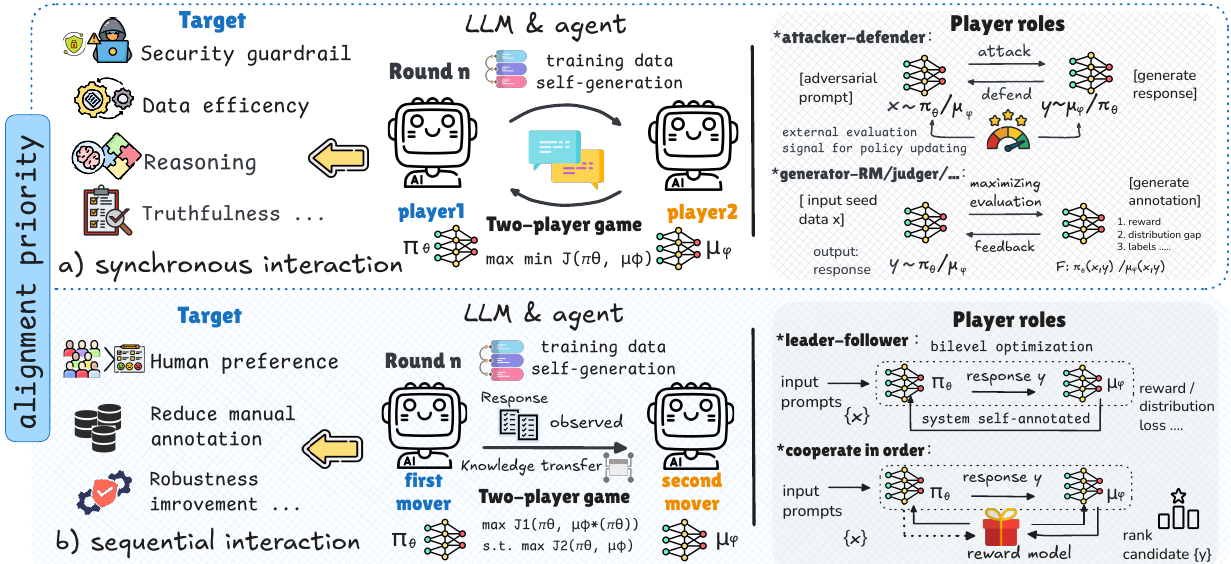}
    \caption{Synchronous interaction for alignment: simultaneous play, adversarial roles, and alignment targets. Sequential interaction for alignment: ordered roles and commitment-based optimization.}
    \label{fig:interaction}
\end{figure*}

\subsubsection{Synchronous Interaction}
When interaction is synchronous, both sides act within the same round, and alignment is improved through repeated exposure to hard cases generated during play, as shown in Fig.~\ref{fig:interaction}(a).

\paragraph{Attacker-defender} 
This setting has been particularly influential in safety alignment. GPO~\cite{zheng2025toward} formulates alignment as a max--min game between an attacker and a defender, showing that adversarial prompt generation can uncover vulnerabilities while improving robustness through iterative updates. Closely related work applies the same intuition to toxicity mitigation, guardrail training, and jailbreak robustness, either by training separate attacker--defender models or by alternating these roles within a single model~\cite{jatova2024employing,deng2026enhancing,liu2026chasing,guo2025mtsa}. Across these studies, synchronous play functions less as an abstract game metaphor than as a mechanism for generating stress cases that standard alignment pipelines may miss.

\paragraph{Generator-evaluator.} 
The same interaction pattern has also been repurposed to reduce dependence on human annotation. APO~\cite{cheng2024adversarial} couples the policy with a reward model in an adversarial loop to address the mismatch between evolving model outputs and static preference supervision. SPIN~\cite{chen2024self} instead treats self-improvement as iterative self-distillation, while LANA~\cite{azarafrooz2024language} further shows that self-rewarding can remain effective even under noisy or suboptimal demonstrations. What connects these methods is that interaction does not merely refine the policy; it also becomes a means of producing or filtering the supervision signal itself.

This interactive template has further extended to reasoning and truthfulness. SPC~\cite{chen2025spc} uses adversarial self-play between a generator and critic to improve step-level reasoning and support test-time search, whereas SPAG~\cite{cheng2024self} relies on adversarial language games to strengthen reasoning-oriented dialogue. PEG~\cite{chen2025incentivizing} pushes the idea toward truthful evaluation by casting peer assessment as a game in which truthful reporting is incentivized. Taken together, these studies suggest that synchronous interaction is most useful when alignment requires continual pressure from challenging inputs, self-generated supervision, or competing evaluations. Its limitations, however, remain clear: many methods are computationally intensive, sensitive to training dynamics, and still rely on assumptions that are difficult to satisfy in realistic black-box settings.

\subsubsection{Sequential Interaction}

\paragraph{Leader-follower.} 
Once the order of moves becomes explicit, interaction changes character. The earlier player can commit to a strategy that reshapes the response of the later player in Fig.~\ref{fig:interaction}(b), making this setting closer to bilevel or Stackelberg optimization than to simultaneous self-play. This perspective has been especially useful for preference learning. STA-RLHF~\cite{makar2024sta} models the language model and reward model as leader and follower, respectively, and searches for a Stackelberg equilibrium under nested optimization. SPAC~\cite{ji2024self} preserves the same leader--follower structure but replaces multi-stage updates with a single-timescale procedure, yielding stronger practical appeal in offline settings with sparse preference data.

\paragraph{Orderly cooperation.} 
A similar logic also appears when sequential interaction is used to improve data quality and robustness. SGPO~\cite{chu2025stackelberg} studies alignment under scarce and noisy preferences through a Stackelberg-style formulation with regret guarantees under bounded distribution shift. Anyprefer~\cite{zhou2025anyprefer} combines automatic synthesis, external tools, and a cooperative Markov game to mitigate self-reward bias while increasing sample diversity. CORY~\cite{ma2024coevolving} goes a step further by using sequential cooperative multi-agent reinforcement learning and role exchange to stabilize fine-tuning. In these settings, ordered interaction is valuable not only because it captures asymmetric roles more faithfully, but also because it offers a natural way to control how evaluators, generators, and auxiliary tools influence one another over time.

These interaction-based approaches move alignment beyond static supervision, but they also overlook a problem: alignment goals, evaluators, and environments may all evolve over time. We therefore turn next to dynamic evolution in alignment.

\subsection{Challenge Three: Temporal Dynamics}
\label{sec:challenge_3}
The previous two sections focus on preference structure and interaction structure at a given stage of learning or deployment. We now turn to a third aspect that cuts across both: temporal change. Once systems interact repeatedly with humans, other agents, and changing environments, alignment can no longer be treated as a one-shot objective fixed at training time. Instead, the central challenge is to maintain behavioral consistency as preferences, incentives, and external conditions evolve. In this sense, dynamic evolution is not a separate alignment category so much as a temporal extension of the preceding discussion. This has motivated a growing body of work on alignment through adaptive evolution and co-evolution.

\subsubsection{Adaptive Evolution}
\begin{figure*}[t!]
    \centering
    \includegraphics[width=0.89\linewidth]{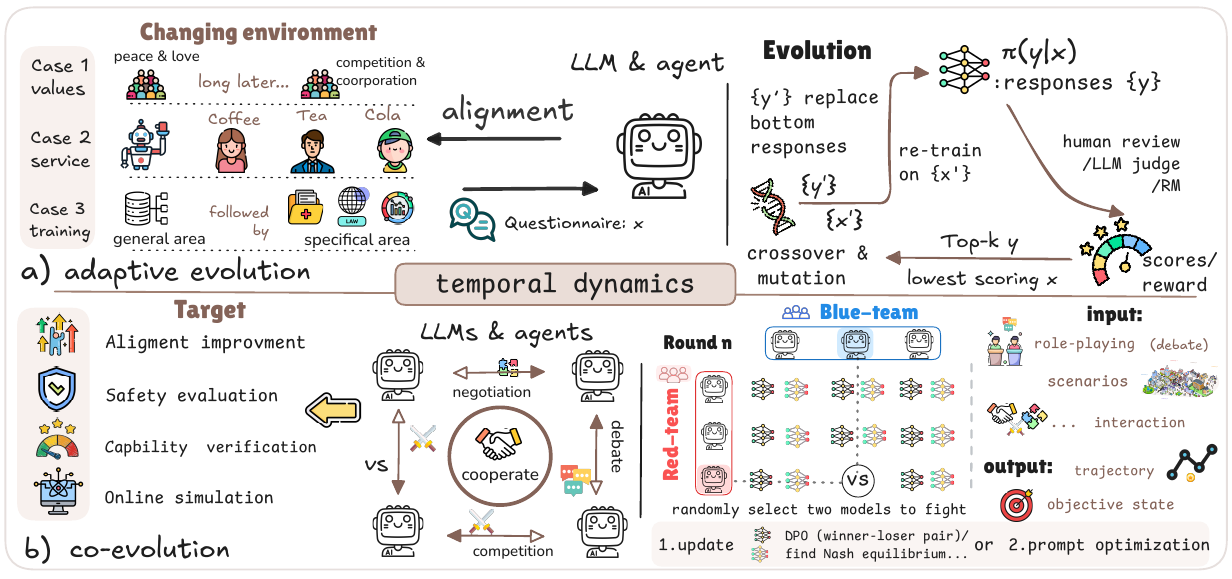}
    \caption{Adaptive evolution in alignment: changing social environments and a typical update framework. Co-evolution in alignment: multiple agents evolving through competition, cooperation, and social interaction.}
    \label{fig:evolution}
\end{figure*}

Adaptive evolution focuses on how AI systems update under temporally varying social environments, as illustrated in Fig.~\ref{fig:evolution}(a).

\paragraph{Trajectory-based evolution.} 
 One line of work formulates this problem through sequential decision-making or trajectory-based evaluation. ProgressGym~\cite{qiu2024progressgym}, for example, frames alignment as moral progress over time and studies how historical moral data can guide future decisions. Relatedly, Klassen et al.~\cite{klassen2024pluralistic} propose a temporal pluralistic alignment framework that evaluates party satisfaction over trajectories rather than isolated outputs, allowing alignment to depend on long-horizon and non-Markovian requirements.

\paragraph{Selection-based evolution.}
A complementary line introduces evolutionary mechanisms such as mutation, selection, and environmental variation. Li et al.~\cite{li2024agent} embed social norms into a genetic process so that agents adapt through repeated selection under changing group behavior, while Suzuki and Arita~\cite{suzuki2024evolutionary} use evolutionary games to study the emergence of cooperative traits in LLM-based agents. Ye et al.~\cite{ye2024evolving} further model prompt mutation as a way to simulate novel environments and formulate alignment as a creator--solver game. Across these studies, the common intuition is that alignment should remain responsive to gradual changes in norms, tasks, and social expectations, rather than being anchored to a static preference snapshot.

\subsubsection{Co-evolution}
Adaptive evolution emphasizes how a model or agent tracks environmental change, while co-evolution focuses on settings in which the environment itself is partly generated by other evolving agents, as shown in Fig.~\ref{fig:evolution}(b).

\paragraph{Game-theoretic algorithm modeling.} 
This perspective is particularly useful when alignment depends on multi-agent interaction, population diversity, or open-ended strategic adaptation. SPARTA~\cite{jiang2025sparta} studies collective alignment through repeated combat, evaluation, and learning among multiple models, while Ma et al.~\cite{ma2026evolving} cast safety alignment as a red--blue team game in which attack and defense strategies evolve together. These approaches suggest that robustness may emerge not only from stronger objectives, but also from continued exposure to adaptive opponents and evaluators.

\paragraph{Scenario simulation.} The same co-evolutionary perspective also underlies a broader set of social simulation environments. Recent work uses agent societies, role-playing communities, and sandbox worlds to study how norms, cooperation, and governance emerge through interaction~\cite{liu2024training,lai2024position,li2023camel,dai2026artificial}. In parallel, strategically structured environments such as Alympics~\cite{mao2025alympics} and FairMindSim~\cite{lei2026llms} are used to probe bargaining, fairness, and collective behavior under resource constraints or ethical dilemmas. Debate-based methods extend this logic to inference-time improvement: by allowing multiple agents to critique and challenge one another, they can improve reasoning and factuality without retraining~\cite{liang2024encouraging,du2024improving}. Together, these studies broaden alignment from single-model optimization to the analysis of collective behavior in evolving social systems, but many of these environments remain behaviorally rich while still offering limited visibility into the internal mechanisms that drive adaptation. 
\section{Future Work Directions}
\label{sec:future_work}

\textbf{Fairness guarantees}
Pluralistic societies typically do not rely on a single value standard, so alignment should not only search for a `true' value objective but also design fair procedures for deciding which values guide model behavior~\cite{macintyre2013after}. Existing multi-objective methods introduce fairness constraints, such as demographic parity, equalized odds, group-DRO, and worst-group lower bounds~\cite{agnihotri2026multi,chakraborty2024maxmin}. However, soft regularization, such as imposing a finite penalty on the objective function, typically gives weak guarantees, while hard constraints can reduce utility or destabilize training~\cite{tran2024effects}. A promising direction is to model fairness-utility trade-offs as games with social-welfare objectives~\cite{chen2024assessing}, including Nash welfare, Leximin~\cite{barman2026compatibility}, and Pareto-optimal solution families~\cite{censor1977pareto}. Future work should further combine adaptive weighting with procedural safeguards from Constitutional AI, social choice, voting, and public participation, so that both outcomes and decision procedures remain legitimate and adaptable~\cite{ZhaoWP24,huang2024collective}.

\textbf{Continual alignment under drift.}
Online tool use, continual preference learning, and scenario migration can update useful capabilities while gradually overwriting aligned behavior. This causes catastrophic forgetting and value drift, especially in adversarial testing and long multi-turn interaction. From a game-theoretic view, alignment maintenance can be modeled as a dynamic interaction among the model, deployment environment, and strategic auditors or adversaries, where feedback streams may expose or induce degradation. Existing remedies include continual or online RLHF~\cite{zhang2024cppo}, trust-region or strongly regularized updates~\cite{li2024revisiting}, and generative replay under data-access limits~\cite{huang2024mitigating}. However, most methods optimize general retention rather than treating alignment-critical knowledge as the main retention target~\cite{wang2024comprehensive}. Future work should combine continual learning with hierarchical or retrieval-augmented memory~\cite{wang2023augmenting,borgeaud2022improving}, classical anti-forgetting mechanisms~\cite{chaudhry2019continual,kirkpatrick2017overcoming}, and adversarially robust updates~\cite{khan2022adversarially,mirzadeh2020understanding}, so that safety-relevant knowledge remains stable under changing or manipulated data flows.

\textbf{Visual simulation for explainable AI.}
AI alignment remains difficult to inspect when simulations only report external behavior. Controlled social simulations, agent societies, and policy games, as discussed in Sec.~\ref{sec:challenge_3}, can reveal dynamic failures in interaction, adaptation, and social feedback. Yet many such environments remain behaviorally rich but mechanistically shallow. Future work should connect these environments with mechanistic interpretability, rule-based dynamic models, intervention logging, and counterfactual analysis. This would turn simulation from outcome-level evaluation into process-level diagnosis, making long-term alignment easier to inspect, explain, and steer.

\section{Conclusion }
\label{sec:conclusion}
We surveyed AI alignment through a game-theoretic lens, organizing existing work around preference diversity, alignment priority, and temporal dynamics. This perspective highlights how strategic interactions can help formalize conflicting preferences, interactive alignment, and evolving values, while also revealing where current theoretical guarantees remain limited. We hope this framework provides a foundation for developing more robust, adaptive, and verifiable alignment methods.


\newpage
\section*{Limitations}

This survey focuses primarily on post-training alignment work for LLMs and LLM-based agents that can be meaningfully analyzed through game-theoretic concepts. It therefore does not aim to exhaustively cover adjacent literature on pre-training objectives, inference-time defenses, mechanistic interpretability, or governance frameworks that do not directly adopt this lens. In addition, the reviewed areas are uneven in theoretical maturity: preference-learning work currently admits stronger formal guarantees than many studies on social interaction, simulation, and long-term evolution. As a result, some parts of the survey necessarily provide a looser synthesis than others.

\section*{Ethics Statement}
While game-theoretic frameworks can help clarify incentives and interaction structures in alignment, they may also simplify complex moral, cultural, and institutional factors. Therefore, the methods reviewed in this paper should not be regarded as complete solutions to alignment or as substitutes for human oversight and governance.  We encourage readers to interpret the reviewed approaches with caution, especially in safety-critical or socially sensitive settings.

\bibliography{custom}

\newpage
\appendix
\section{Supplementary Analysis and Tables}\label{appx}

This appendix provides supplementary material for the game-theoretic alignment frameworks discussed in the main text. It consists of two parts. 
First, we provide detailed tabular comparisons of representative alignment methods corresponding to the three branches of the taxonomy introduced in Sec.~\ref{sec:alignment_methods}: preference diversity, alignment priority, and temporal dynamics. These tables complement the main-text discussion by exposing methodological details such as game formalization, convergence properties, self-play, feedback sources, and online optimization that cannot be fully presented in the main text due to space limitations. 
Second, we provide a supplementary discussion of game-based evaluation, together with representative benchmarks for assessing the strategic and social behaviors of LLMs and LLM-based agents.

\subsection{Detailed Comparison of Game-Theoretic Alignment Methods}\label{appx:details}

While representative methods are discussed in the main text,  Tabs.~\ref{tab:alignment_framework_pref} to \ref{tab:alignment_framework_temp} provide a more fine-grained comparison of their formulations and methodological properties, such as equilibrium convergence, self-game, and feedback. These tables correspond to the three branches of the taxonomy introduced in Sec.~\ref{sec:alignment_methods}. Specifically, Tab.~\ref{tab:alignment_framework_pref} summarizes methods from the perspective of \emph{preference diversity}, Tab.~\ref{tab:alignment_framework_priority} focuses on \emph{alignment priority and interaction structure}, and Tab.~\ref{tab:alignment_framework_temp} summarizes methods characterized by \emph{temporal dynamics}.

\subsection{Game-Based Evaluation and Benchmarks}\label{appx:benchmark}

Beyond serving as a modeling and optimization tool for alignment, game theory also provides controlled environments for evaluating the strategic and social behaviors of LLMs and LLM-based agents. The interactive nature of games makes it possible to examine properties that are difficult to capture through static language-generation benchmarks, including strategic reasoning, adaptation, cooperation, competition, negotiation, trust, and coordination. This evaluation perspective is complementary to the alignment frameworks discussed in the main text. Rather than directly optimizing an alignment objective, game-based evaluation treats strategic interactions as test environments in which model behavior can be systematically observed and quantified. We summarize representative studies under two broad paradigms: (1) evaluation based on classical games and (2) evaluation based on social games. A structured comparison of representative benchmarks, evaluation targets, game formulations, and metrics is provided in Tab.~\ref{tab:eval_benchmark}.

\begin{table*}[t!]
\centering
\caption{Overview of alignment frameworks based on preference diversity.}
\label{tab:alignment_framework_pref}

\begingroup
\scriptsize
\setlength{\tabcolsep}{4pt}
\renewcommand{\arraystretch}{1.08}
\setlength{\extrarowheight}{1pt}
\begin{adjustbox}{max width=\linewidth}
\begin{tabular}{
    >{\raggedright\arraybackslash}p{2.35cm}
    >{\raggedright\arraybackslash}p{3.05cm}
    >{\raggedright\arraybackslash}p{1.75cm}
    >{\raggedright\arraybackslash}p{4.25cm}
    >{\raggedright\arraybackslash}p{3.45cm}
    c c c c
}
\toprule
\multicolumn{1}{c}{} & \multicolumn{8}{c}{\textbf{Preference alignment}} \\
\cmidrule(lr){2-9}
\rowcolor{headergray}
\textbf{Direction}& \textbf{Framework} & \textbf{Regularization}& \textbf{Formalization} & \textbf{Convergence} & \textbf{\makecell{Preference\\modeling}} & \textbf{Self-play?} &\textbf{\makecell{Robust\\alignment?}}& \textbf{Online?}\\
\midrule

\multirow{14}{*}{\textbf{\makecell[l]{General\\preference}}}
& IPO\cite{azar2024general} & Explicit & Best response (fixed strategy) & Average-iterate ($\epsilon$-OPT) & \symempty & \symempty & \symempty & \symempty\\

& NLHF\cite{munos2024nash} & Explicit & Two-player game (constant-sum) & Last-iterate (NE) & \symfilled & \symhybrid & \symempty & \symfilled\\

& IPO-MD\cite{calandriello2024human} & Explicit & Two-player game & Exist (NE) & \symempty & \symhybrid & \symempty & \symfilled\\ 

& Iterative RLHF\cite{ye2024online} & Explicit & Two-player game (mini-max) & Exist ($\epsilon$-NE) & \symfilled & \symfilled & \symempty & \symfilled\\

& SPO\cite{swamy2024minimaximalist} & - & Two-player game (zero-sum) & Average-iterate ($\epsilon$-Minimum Winner) & \symfilled & \symfilled & \symempty & \symfilled\\

& DNO\cite{rosset2024direct} & - & Two-player game (zero-sum) & Average-iterate (NE) & \symfilled & \symfilled & \symempty & \symhybrid \\

& REBEL\cite{gao2024rebel} & - & Two-player game (zero-sum) &Average-iterate (\(\epsilon\)-OPT) & \symfilled & \symfilled & \symempty & \symhybrid \\

& SPPO\cite{wu2025self} & - & Two-player game (constant-sum) & Average-iterate ($\epsilon$-NE) & \symfilled & \symfilled & \symempty & \symhybrid\\

& TANPO\cite{wangprovably} & Explicit & Two-player game (zero-sum) & Average-iterate ($\epsilon$-NE) & \symempty & \symfilled & \symempty & \symfilled\\

& ONPO\cite{zhang2025improving} & - & two-player game (zero-sum) & Average-iterate ($\epsilon$-NE) & \symempty & \symfilled & \symempty & \symfilled\\

& INPO\cite{zhangiterative} & Explicit & Two-player game & Last-iterate ($\epsilon$-NE) & \symempty & \symfilled & \symempty & \symfilled\\

& COMAL\cite{liu2026comal} & - & Two-player game (zero-sum) & Last-iterate (NE) & \symfilled & \symempty & \symfilled & \symfilled\\

& MPO\cite{wang2025magnetic} & - & Two-player game (constant-sum) & Last-iterate (NE) & \symfilled & \symfilled & \symempty & \symfilled\\

& RSPO~\cite{tang2025game} & Explicit & Two-player game & Last-iterate (NE) & \symfilled & \symfilled & \symempty & \symfilled \\

\midrule

\multirow{15}{*}{\textbf{\makecell[l]{Heterogeneous\\preference}}}

& \cite{chidambaram2026direct} & Explicit & Social welfare optimization & Average-iterate ($\epsilon$-OPT) & \symfilled & \symempty & \symfilled & \symempty\\

& \cite{kim2026beyond} & Explicit & Social welfare optimization & - & \symfilled & \symempty & \symempty & \symempty\\

& MOP\cite{xiongprojection} & Explicit & Social welfare optimization & Average-iterate (OPT) & \symfilled & \symempty & \symempty & \symhybrid\\

& Pessimistic Nash Bargaining\cite{zhong2024provable} & Implicit & Social welfare optimization & Exist ($\epsilon$-OPT) & \symfilled & \symempty & \symempty & \symempty\\

& GRPO\cite{ramesh2024group} & Implicit & Social welfare optimization & Average-iterate ($\epsilon$-OPT) & \symfilled & \symempty & \symfilled & \symempty\\

& CLP~\cite{wang2024conditional} & Explicit & Multi-objective optimization & Exist ($\epsilon$-OPT) & \symfilled & \symempty & \symempty & \symfilled\\

& Panacea\cite{zhong2024panacea} & Both & Multi-objective optimization & Exist (Pareto-OPT) & \symfilled & \symempty & \symempty & \symfilled\\

& MO-ODPO\cite{gupta2025robust} & Explicit & Multi-objective optimization & - & \symfilled & \symempty & \symempty & \symfilled\\

& SIPO\cite{li2025self} & Explicit & Multi-objective optimization & - & \symfilled & \symempty & \symempty & \symhybrid\\

& DPA\cite{wang2024arithmetic} & - & Multi-objective optimization & - & \symfilled & \symempty & \symempty & \symhybrid\\

& DRM\cite{luo2025rethinking} & Explicit & Multi-objective optimization & - & \symfilled & \symempty & \symempty & \symempty\\

& RLHF~Game\cite{sun2025mechanism} & Explicit & Multi-player game (mechanism design) & - & \symfilled & \symempty & \symempty & \symempty\\

& Dynamic~Bayesian~Game\cite{hao2025online} & Explicit & Multi-player game (dynamic Bayesian) & Average-iterate ($\epsilon$-OPT) & \symfilled & \symempty & \symempty & \symfilled\\

& BIG\cite{wu2025battling} & - & Multi-player game (cooperative) & Last-iterate (NE) & \symfilled & \symempty & \symempty & \symempty \\

& $\langle M,N,\epsilon,\delta\rangle$-Agreement\cite{nayebi2026intrinsic} & - & Multi-player game (theoretical) & - & \symempty & \symempty & \symempty & \symfilled\\

\bottomrule
\end{tabular}
\end{adjustbox}

\parbox{0.98\linewidth}{\footnotesize \textbf{Notes.} (1) \emph{Regularization}: explicit penalties or procedures that constrain the model's policy, learning, and outputs. (2) \emph{Formalization}: how the alignment problem is modeled. (3) \emph{Convergence}: based on theoretical proof or experimental observation, the optimal strategy exists when converging to an equilibrium point, or explicitly reaching a consensus state in decision-making; some methods provide stronger guarantees such as average-iterate and last-iterate convergence. (4) \emph{Preference modeling}: a preference probability model (reward model) is learned from human preference data for candidate outputs. (5) \emph{Self-play}: the same model or homologous variants/clones engage in mutual competition, collaboration, or evaluation within a task or game. (6) \emph{Robust alignment}: a Nash equilibrium policy guarantees at least a $50\%$ win (or preference) rate against any alternative policy under general preferences, and guarantees the worst-group performance under heterogeneous preferences. (7) \emph{Online}: new training data are obtained from the current model and the parameters are updated accordingly. Symbols {\protect\symfilled}, {\protect\symempty}, and {\protect\symhybrid} denote conditions that are met, not met, and adjustable mixed settings, respectively; `-' indicates not applicable or not reported.}
\endgroup
\end{table*}

\begin{table*}[t!]
\centering  
\caption{Overview of alignment frameworks based on alignment priority.}
\label{tab:alignment_framework_priority}

\begingroup
\scriptsize
\setlength{\tabcolsep}{4pt}
\renewcommand{\arraystretch}{1.08}
\setlength{\extrarowheight}{1pt}
\begin{adjustbox}{max width=\linewidth}
\begin{tabular}{
    >{\raggedright\arraybackslash}p{2.45cm}
    >{\raggedright\arraybackslash}p{3.10cm}
    >{\raggedright\arraybackslash}p{2.30cm}
    >{\raggedright\arraybackslash}p{3.25cm}
    >{\raggedright\arraybackslash}p{3.55cm}
    c c c c
}
\toprule
\multicolumn{1}{c}{} & \multicolumn{8}{c}{\textbf{Enhancement via interaction}} \\
\cmidrule(lr){2-9}
\rowcolor{headergray}
\textbf{Direction}& \textbf{Framework} & \textbf{Target} & \textbf{Formalization} & \textbf{Interaction structure} & \textbf{Convergence} & \textbf{\makecell{External\\feedback}} & \textbf{Self-play?} & \textbf{Online?}\\
\midrule

\multirow{11}{*}{\textbf{\makecell[l]{Synchronous\\interaction}}}
& \cite{jatova2024employing} & Safety alignment & Two-player game & LLM vs adversarial generator & Exist (NE) & \symfilled & \symempty & \symfilled \\

& GPO~\cite{zheng2025toward} & Safety alignment & Two-player game (zero-sum) & LLMs (attacker vs defender) & Average-iterate (NE) & \symfilled & \symempty & \symfilled \\

& SELF-REDTEAM~\cite{liu2026chasing} & Safety alignment & Two-player game (zero-sum) & LLM (attacker / defender) & - & \symfilled & \symfilled & \symfilled \\

& MTSA~\cite{guo2025mtsa} & Safety alignment & Multi-turn RL & LLMs (attacker vs defender) & - & \symfilled & \symempty & \symhybrid \\

& SPAG~\cite{cheng2024self} & Reasoning & Two-player game (zero-sum) & LLM (attacker / defender) & - & \symfilled & \symfilled & \symempty \\

& APO~\cite{cheng2024adversarial} & Effective alignment & Two-player game (mini-max) & LLM vs RM & - & \symempty & \symempty & \symfilled \\

& SPIN~\cite{chen2024self} & Self-improvement & Two-player game & LLM (generator / distinguisher) & Exist & \symempty & \symfilled & \symhybrid \\

& LANA~\cite{azarafrooz2024language} & Self-alignment & Two-player game (zero-sum) & LLM (generation / evaluation) & Average-iterate ($\epsilon$-NE) & \symempty & \symfilled & \symfilled \\

& DuoGuard~\cite{deng2026enhancing} & Safety alignment & Two-player game (mini-max) & LLMs (generator vs classifier) & Last-iterate (NE) & \symempty & \symempty & \symhybrid \\

& SPC~\cite{chen2025spc} & Reasoning & Two-player game & LLM (generator / critic) & - & \symempty & \symfilled & \symfilled \\

& PEG~\cite{chen2025incentivizing} & Truthfulness & Peer elicitation game & LLMs (peer discriminators) & Last-iterate (Truthful NE) & \symempty & \symempty & \symfilled \\

\midrule

\multirow{5}{*}{\textbf{\makecell[l]{Sequential\\interaction}}}
& STA\textendash RLHF~\cite{makar2024sta} & Human preference & Two-player game (Stackelberg) & LM (leader) vs RM (follower) & - & \symempty & \symempty & \symhybrid \\

& SPAC~\cite{ji2024self} & Human preference & Two-player game (Stackelberg) & LLM (policy model / critic) & Average-iterate ($\epsilon$-OPT) & \symempty & \symfilled & \symempty \\

& SGPO~\cite{chu2025stackelberg} & Effective alignment & Two-player game (Stackelberg) & LLM vs pref. distribution & Last-iterate (SE) & \symempty & \symempty & \symhybrid \\

& Anyprefer~\cite{zhou2025anyprefer} & Effective alignment & Two-player game (Markov) & LLMs (target model \& judger) & - & \symfilled & \symempty & \symempty \\

& CORY~\cite{ma2024coevolving} & Robustness & Cooperative multi-agent RL & LLM (pioneer / observer) & - & \symfilled & \symfilled & \symfilled \\

\bottomrule
\end{tabular}
\end{adjustbox}

\parbox{0.98\linewidth}{\footnotesize \textbf{Notes.} (1) \emph{Target}: research question. (2) \emph{Formalization}: how the alignment problem is modeled. (3) \emph{Interaction structure}: players and roles engaging in interaction. (4) \emph{Convergence}: based on theoretical proof or experimental observation, the optimal strategy exists when converging to an equilibrium point, or explicitly reaching a consensus state in decision-making; some methods provide stronger guarantees such as average-iterate and last-iterate convergence. (5) \emph{External feedback}: the AI model (agent) relies on externally added, manually annotated data for optimization and updates, rather than self-generating new annotations via interaction. (6) \emph{Self-play}: the same model or homologous variants/clones engage in mutual competition, collaboration, or evaluation within a task or game. (7) \emph{Online}: new training data are obtained from the current model and the parameters are updated accordingly. Symbols {\protect\symfilled}, {\protect\symempty}, and {\protect\symhybrid} denote conditions that are met, not met, and adjustable mixed settings, respectively; `-' indicates not applicable or not reported.}
\endgroup
\end{table*}

\begin{table*}[t!]
\centering
\caption{Overview of alignment (simulation) frameworks based on temporal dynamics.}
\label{tab:alignment_framework_temp}
\begingroup
\scriptsize
\setlength{\tabcolsep}{4pt}
\renewcommand{\arraystretch}{1.08}
\setlength{\extrarowheight}{1pt}
\begin{adjustbox}{max width=\linewidth}
\begin{tabular}{
    >{\raggedright\arraybackslash}p{2.50cm}
    >{\raggedright\arraybackslash}p{3.10cm}
    >{\raggedright\arraybackslash}p{2.80cm}
    >{\raggedright\arraybackslash}p{3.40cm}
    >{\raggedright\arraybackslash}p{2.55cm}
    >{\raggedright\arraybackslash}p{2.15cm}
    c c
}
\toprule
\multicolumn{1}{c}{} & \multicolumn{7}{c}{\textbf{Dynamic evolution}} \\
\cmidrule(lr){2-8}
\rowcolor{headergray}
\textbf{Direction} & \textbf{Framework} & \textbf{Target} & \textbf{Formalization} & \textbf{Application} & \textbf{Stage} & \textbf{Convergence} & \textbf{Online?}\\
\midrule

\multirow{5}{*}{\textbf{\makecell[l]{Adaptive\\evolution}}}
& \cite{klassen2024pluralistic} & Temporal pluralism values & Non-Markovian process & Measurement & - & - & -\\

& \cite{suzuki2024evolutionary} & Social cooperation & Evolutionary game (mutation) & Simulation & Inference & - & - \\

& ProgressGym~\cite{qiu2024progressgym} & Social moral progress & Partially observable MDP & Improvement & Post-training & - & -\\

& Evolutionary Agent~\cite{li2024agent} & Norms and cooperation & Evolutionary game (mutation) & Improvement & Inference & - & - \\

& EVA~\cite{ye2024evolving} & Preference alignment & Creator-solver game (mutation) & Improvement & Post-training & - & \symfilled\\

\midrule

\multirow{10}{*}{\textbf{Co-evolution}}
& SPARTA~\cite{jiang2025sparta} & Preference / instruction & Collective competition & Improvement & Post-training & - & \symhybrid\\

& Red-Team Game~\cite{ma2026evolving} & Safety alignment & Red-blue team game & Improvement & Post-training & Exist ($\epsilon$-NE) & \symfilled\\

& SSI~\cite{liu2024training} & Social alignment & Social interaction & Improvement & Post-training & Exist & \symempty\\

& Artificial Leviathan~\cite{dai2026artificial} & Social order & Social interaction (survival sandbox) & Simulation & Inference & - & -\\

& CAMEL~\cite{li2023camel} & Social cooperation & Role-playing interaction (dialogue) & Simulation & Inference & - & -\\

& AI Collectives~\cite{lai2024position} & Social alignment & Social interaction (community) & Simulation & Inference & - & -\\

& Alympics~\cite{mao2025alympics} & Empirical game theory & Strategic games (survival resources) & Simulation & Inference & - & -\\

& FairMindSim~\cite{lei2026llms} & Social value alignment & Ethical dilemma & Simulation & Inference & - & -\\

& \cite{du2024improving} & Reasoning / factuality & Multi-agent debate & Improvement & Inference & Exist (Consensus) & -\\

& MAD~\cite{liang2024encouraging} & Reasoning & Multi-agent debate & Improvement & Inference & - & -\\

\bottomrule
\end{tabular}
\end{adjustbox}

\parbox{0.98\linewidth}{\footnotesize \textbf{Notes.} (1) \emph{Target}: research question. (2) \emph{Formalization}: how the process of dynamic evolution is driven or modeled, such as game, social interaction, and debate. (3) \emph{Application}: quantify and enhance the alignment performance of AI systems, or conduct interdisciplinary research by simulating social behavior and group evolution processes through AI agents, such as social sciences and empirical game theory. (4) \emph{Stage}: the phase during which the training or evaluation framework is running. (5) \emph{Convergence}: based on theoretical proof or experimental observation, the optimal strategy exists when converging to an equilibrium point, or explicitly reaching a consensus state in decision-making. (6) \emph{Online}: new training data are obtained from the current model and the parameters are updated accordingly. Symbols {\protect\symfilled}, {\protect\symempty}, and {\protect\symhybrid} denote conditions that are met, not met, and adjustable mixed settings, respectively; `-' indicates not applicable or not reported.}
\endgroup
\end{table*}

\subsubsection{Evaluation Based on Classic Games}

Classic games, such as 2$\times$2 matrix games~\cite{herr2024large} and Rock-Paper-Scissors~\cite{fan2024can}, are typically used to characterize capability performance of LLMs and LLM-based agents by employing strategic interaction and reasoning tasks. This capability verification aims to transform the risks and protection requirements exposed by external behavioral assessments into executable capability constraints and verification protocols, while specifying response boundaries and the system's expected performance under normal conditions~\cite{sun2025game}. Related advances are summarized into two categories.

One direction explores cognitive intelligence (CI), which typically focuses not only on the quality of language generation and knowledge retrieval~\cite{niu2024large, wang2025bring}, but also encompasses strategic reasoning~\cite{li2025system}, planning, and robust decision-making under adversarial structures~\cite{ni2025survey, mirzadeh2025gsmsymbolic, chen2024benchmarking}. In the context of LLMs and LLM-based agents evaluation, a representative line of work treats LLMs and LLM-based agents as decision makers in controlled strategic tasks, and uses the resulting interaction traces to audit whether the model exhibits game-theoretic rationality and stable strategic reasoning rather than surface-level pattern matching. For example, Fan et al.~\cite{fan2024can} map rationality onto three dimensions: preference formation, belief adjustment, and optimal action. They then systematically assess the rationality of large language models across three classic games. Further, Wang et al.~\cite{wang2024tmgbench} extend the coverage of game types and systematically evaluate the accuracy of strategic reasoning, based on a 2×2 game topology system. Meanwhile, another studies examined factors influencing the robustness of strategic reasoning in matrix games~\cite{lore2024strategic} and Grid-Based Game~\cite{topsakal2024evaluating}. These investigations ultimately revealed that strategic choices are affected by positional presentation, payoff structures, narrative contexts, and prompt format. For multiplayer competitive environments, Lu~\cite{lu2024strategic}  evaluated the efficient inference level of the LLM-agent through a beauty contest and observed convergence behavior in repeated games.

Additionally, some classic games and their more complex extensions are also used to evaluate social intelligence (SI) of AI systems in interactive decision-making, including various competence such as cooperation and competition~\cite{sreedhar2025simulating}. Specifically, the Prisoner's Dilemma and repeated games are widely used as standard test platforms to compare the stability and differences of LLMs in terms of cooperative, punitive, and other behaviors in social dilemmas~\cite{akata2025playing, fontana2025nicer}. Meanwhile, some studies emphasize behaviors that reflect social preferences and norms, such as trust and fairness,  and use Trust games~\cite{xie2024can} and Ultimatum~\cite{zakazov2024assessing} to demonstrate  models' consistency with human behavior. In addition to the social competence mentioned above, SI also emphasizes interpersonal abilities, which are typically validated using classical communication games. Regarding this, Davidson et al.~\cite{davidson2024evaluating} proposed using multi-round negotiation as an evaluation framework to simultaneously measure the negotiation and alignment behavior of language models under self-play and cross-play conditions, and pointed out that even strong models will experience policy adaptation and stability issues when the opponent changes. Similarly, Xia et al.~\cite{xia2024measuring} formalized bargaining as an asymmetric incomplete information game, constructed a quantitative benchmark based on real price datasets, and evaluated the models for both buyers and sellers separately.

\subsubsection{Evaluation based on Social games}

Compared with classic games, social games offer a more comprehensive social process by introducing role-playing, norm formation, information dissemination, and group decision-making, thus more closely approximates real-world deployment scenarios for system-level supervision and evaluation. Meanwhile, research based on this type of game theory often designs more flexible social interaction and narrative scenarios, providing a wealth of benchmarks as one of its contributions. Moreover, when a LLM acts as a social player, this game transforms language itself into part of the action space, making it easier to expose the model's true strategic capabilities and failure modes. Here, we need to emphasize that social games differs from the scenario simulation in dynamic evolution mentioned above. The difference is reflected in the fact that evaluation studies use social games as a testing platform to quantify strategy consistency or strategic performance under controlled interactions, while simulation studies use social games as a world model to study the long-term dynamics, emergent norms, and feedback loops between LLMs (agents) and their social environment. Therefore, we will discuss related works separately in this section.

Game evaluation driven by controllable scripts and social roles is gaining popularity among researchers who use language to standardize social competence into reproducible benchmarks and metrics like goal achievement, enabling cross-model comparisons. For example, Zhou et al.~\cite{zhou2025socialeval} propose a scripted social intelligence benchmark SOCIALEVAL, and unify outcome-oriented and process-oriented approaches into the same evaluation paradigm, providing an external criterion for alignment in social interaction. Similarly, Zhou et al.~\cite{zhou2024sotopia} construct an open social interaction environment, using goal-driven social scenarios to enable agents to interact in multiple rounds, thereby evaluating their social intelligence and interaction quality. Meanwhile, some studies focus on role-playing and multi-turn dialogues, using a process-oriented approach to assess whether the model's behavior consistently follows the role setting~\cite{liu2024roleagent, wang2024rolellm}.
In addition to the efforts mentioned above, compared with the two-player mode commonly found in classic games, social games provide more opportunities for interactive research on multi-agent systems via scenarios simulation, within the aspects of collaboration, competition, negotiation, and social reasoning. For example, the benchmark MultiAgentBench~\cite{zhu2025multiagentbench} evaluate multi-agent systems through multi-scenario interaction tasks, while introducing more granular metrics such as communication, planning, and coordination quality to characterize the structural differences in team collaboration processes and competitive behaviors. Similar works include~\cite{wang2024towards, abdelnabi2024cooperation, xie2026m3} aimed at evaluating the policy decision-making and verifiable execution capabilities of models or agents under information asymmetry and resource constraints. Furthermore, given the potential for strategically uncontrolled behavior in systems, which leads to a cascade of risk effects, the safety assessment of multi-agent behavior and the exploration of alignment risks are receiving increasing attention~\cite{curvo2025traitors, motwani2024secret}.

\begin{table*}[t!]
\centering
\caption{Overview of evaluation benchmarks based on games.}
\label{tab:eval_benchmark}

\begingroup
\scriptsize
\setlength{\tabcolsep}{3.5pt}
\renewcommand{\arraystretch}{1.08}
\setlength{\extrarowheight}{1pt}
\begin{adjustbox}{max width=\textwidth}
\begin{tabular}{
    >{\raggedright\arraybackslash}p{2.00cm}
    >{\raggedright\arraybackslash}p{2.65cm}
    >{\raggedright\arraybackslash}p{2.35cm}
    >{\raggedright\arraybackslash}p{2.75cm}
    >{\raggedright\arraybackslash}p{3.10cm}
    >{\raggedright\arraybackslash}p{1.85cm}
    c
}
\toprule
\multicolumn{1}{c}{} & \multicolumn{6}{c}{\textbf{Evaluation and Benchmark}} \\
\cmidrule(lr){2-7}
\rowcolor{headergray}
\textbf{Paradigm} & \textbf{Ref.} & \textbf{Target} & \textbf{Formalization} & \textbf{Metric} & \textbf{Benchmark} & \textbf{Oriented-O/P}\\
\midrule

\multirow{12}{*}{\textbf{\makecell[l]{Classic\\Games}}}
& Fan et al.\cite{fan2024can} & Rationality & \makecell[l]{Dictator Game\\Ring-network Game} & \makecell[l]{Action Accuracy\\Average Payoff\\Preference Consistency} & - & \symempty\\

& Wang et al.~\cite{wang2024tmgbench} & Strategic reasoning & \makecell[l]{Robinson-Goforth\\Topology of 2$\times$2 Games} & \makecell[l]{Bias Degree\\Perfect Accuracy Rate\\Inconsistency Degree} & Tmgbench & \symhybrid \\

& Lor`e et al.\cite{lore2024strategic} & Strategic behavior & Social Dilemmas Games & \makecell[l]{Cooperation Rate\\Dominance Statistic} & - & \symhybrid \\

& Topsaka et al.~\cite{topsakal2024evaluating} & \makecell[l]{Rule comprehension\\Strategic thinking} & Grid-based Games & \makecell[l]{Win/DQ Rates\\Invalid Moves\\Missed Opportunities} & \ding{52} & \symhybrid \\

& Lu~\cite{lu2024strategic} & \makecell[l]{Strategic levels\\Adaptive learning} & Beauty Contest & \makecell[l]{Strategic Degree\\Convergence\\Average Payoffs} & - & \symempty \\

& Sreedhar et al.~\cite{sreedhar2025simulating} & \makecell[l]{Cooperative prosocial\\behavior} & Public Goods Game & \makecell[l]{Average Contribution\\Emergence Frequency} & - & \symhybrid \\

& Akata et al,~\cite{akata2025playing} & \makecell[l]{Cooperation\\Coordination} & \makecell[l]{Finitely repeated\\2$\times$2 games} & \makecell[l]{Score Ratio\\Defection Rate\\Coordination Rate} & - & \symhybrid \\

& Fontan et al.~\cite{fontana2025nicer} & Cooperative behavior & \makecell[l]{Iterated Prisoner's\\Dilemma} & \makecell[l]{Behavioral Traits\\SFEM Scores} & \ding{52} & \symhybrid \\

& Xie et al.~\cite{xie2024can} & \makecell[l]{Human trust\\behavior} & Trust Game & \makecell[l]{Trust Rate\\Valid Response Rate\\Returned/Sent Ratio} & \ding{52} & \symhybrid \\

& Zakazo et al.~\cite{zakazov2024assessing} & Social alignment & Ultimatum Game & \makecell[l]{Acceptance Rate\\Disobedience Ratio} & - & \symhybrid \\

& Davidson et al.~\cite{davidson2024evaluating} & \makecell[l]{Instruction-following\\Faithfulness} & Structured Negotiation & \makecell[l]{Agreement Rate\\Normalized Payoffs} & LAMEN & \symhybrid \\

& Xia et al.~\cite{xia2024measuring} & Bargaining abilities & Bargaining & \makecell[l]{Deal Rate\\First Bid Ratio\\Sum of Normalized Profits} & AmazonHistoryPrice & \symfilled \\

\midrule

\multirow{10}{*}{\textbf{\makecell[l]{Social\\games}}}
& Zhou et al.~\cite{zhou2025socialeval} & Social intelligence & World tree scenarios & \makecell[l]{Goal Achievement Ratio\\Ability Selection Accuracy} & SocialEval & \symhybrid \\

& Zhou et al.~\cite{zhou2024sotopia} & Social intelligence & Mixed-motive Markov games & \makecell[l]{Goal Completion\\Social Rules} & SOTOPIA-EVAL & \symhybrid \\

& Liu.et al.~\cite{liu2024roleagent} & Role consistency & Interactive dialogue & Self-Knowledge accuracy & RoleAgentBench & \symempty \\

& Wang et al.~\cite{wang2024rolellm} & Role consistency & Role-playing dialogue & \makecell[l]{Win rate\\Average ranking} & RoleBench & \symfilled \\

& Zhu et al.~\cite{zhu2025multiagentbench} & Cooperation \& competition & Mutual/conflicting scenarios & \makecell[l]{Task Score\\Milestone-based KPI} & MultiAgentBench & \symhybrid \\

& Wang et al.~\cite{wang2024towards} & Social intelligence & Social sandbox tasks & \makecell[l]{Success Rate\\Goal Condition SR} & STSS benchmark & \symfilled \\

& Abdelnabi et al.~\cite{abdelnabi2024cooperation} & Negotiation \& security & Multi-agent, multi-issue negotiation & \makecell[l]{Final success\\Collective score\\Score leakage ratio} & \ding{52} & \symhybrid \\

& Xie et al.~\cite{xie2026m3} & Social behavior & Mixed-motive games & \makecell[l]{Overall Score\\Talk--Action Consistency\\Trajectory Analysis} & M3-Bench & \symhybrid \\

& Curvo et al.~\cite{curvo2025traitors} & Deception \& trust & \makecell[l]{Communication utterances\\Voting decisions} & \makecell[l]{Traitor Survival Rate\\Trust Network Stability\\Faithful Correctness Rate} & \ding{52} & \symhybrid \\

& Motwani et al.~\cite{motwani2024secret} & Secret collusion & Covert communication & \makecell[l]{Cipher round-trip accuracy\\Coordination rates} & \ding{52} & \symhybrid \\

\bottomrule
\end{tabular}
\end{adjustbox}

\parbox{0.98\textwidth}{\footnotesize \textbf{Notes.} (1) \emph{Target}: research question. (2) \emph{Formalization}: evaluation scenario or benchmark type. (3) \emph{Metric}: quantifiable evaluation metrics that are typically results- or process-oriented. (4) \emph{Benchmark}: self-named benchmarks are listed explicitly, while unnamed benchmarks are represented by \ding{52}. (5) \emph{Oriented-O/P}: the evaluation paradigm follows outcome-oriented (O: {\protect\symfilled}), process-oriented (P: {\protect\symempty}), or both ({\protect\symhybrid}). The symbol `-' is used in the same way as above.}
\endgroup
\end{table*}

\end{document}